# Interpretable Multi-Instance Learning Enables Early Prediction of Key Molecular Alterations from Routine Flow Cytometry in Acute Myeloid Leukemia

## Authors

*Jonathan Legrand, Univ. Bordeaux, CNRS, Inria, Bordeaux INP, IMB, UMR 5251, F-33400 Talence, France*

*Aguirre Mimoun, CHU Bordeaux, Laboratoire d'Hématologie, F-33000 Bordeaux, France*

*Baudouin Denis de Senneville, Univ. Bordeaux, CNRS, Inria, Bordeaux INP, IMB, UMR 5251, F-33400 Talence, France*

*Audrey Bidet, Laboratoire d'Hématologie Biologique, CHU Bordeaux, Avenue Magellan, Bordeaux, Pessac 33600, France*

*Pierre-Yves Dumas, Univ. Bordeaux, INSERM, BRIC, U1312, CHU Bordeaux, Service d'Hématologie Clinique et de Thérapie Cellulaire, F-33000, Bordeaux, France*

*Christèle Etchegaray, Univ. Bordeaux, CNRS, Inria, Bordeaux INP, IMB, UMR 5251, F-33400 Talence, France*

## Corresponding author

*Jonathan Legrand, Univ. Bordeaux, CNRS, Inria, Bordeaux INP, IMB, UMR 5251, F-33400 Talence, France*

*jonathan.legrand@math.u-bordeaux.fr*

# Abstract

**Background:** Molecular testing for *NPM1* and *FLT3*-ITD mutations guides critical early treatment decisions in acute myeloid leukemia (AML), but results can take weeks, long after these decisions must be made. Flow cytometry, already performed within hours of admission as part of routine care, may carry enough signal to predict these mutations directly, without added cost or delay.

**Methods:** We developed an interpretable multi-instance learning classifier based on a decision tree, in which each patient sample is modeled as a collection of individual cells and mutation status is inferred from cell-level predictions. The model was benchmarked against a random forest trained on clinical variables and a deep convolutional neural network adapted for multi-tube flow cytometry data. Performance was assessed by cross-validation on a discovery cohort of 197 patients and tested on an independent cohort of 161 patients, using the area under the receiver operating characteristic curve (AUROC) and positive predictive value.

**Results:** In cross-validation on the discovery cohort, the MIL model achieved mean AUROCs of 0.96 (SD=0.05) for *NPM1* and 0.86 (SD=0.10) for *FLT3*-ITD, outperforming the clinical baseline and matching deep learning approaches. The model then successfully generalized to the independent test cohort of 161 patients, reaching AUROCs of 0.90 (*NPM1*) and 0.82 (*FLT3*-ITD), with positive predictive values of 0.87 and 0.68, respectively. Cell-level interpretation recovered established immunophenotypic signatures ($CD33^{+}/CD34^{-}$ for *NPM1*-mutated cases, $CD33^{+}$/low side-scatter for *FLT3*-ITD), directly linking model predictions to known biology.

**Conclusions:** These results show that an interpretable model applied to data already collected in routine care can predict AML molecular status within hours, offering a practical route to earlier, biology-informed treatment decisions.

# Introduction

Acute Myeloid Leukemia (AML) is an aggressive cancer characterized by the rapid proliferation of immature blood cells of the myeloid lineage within the bone marrow (1). Diagnosis usually begins with peripheral blood analysis and is confirmed with a bone marrow aspirate. The evaluation of marrow samples includes microscopy to quantify blast cells and assess their morphology, as well as flow cytometry to determine disease lineage and identify antigen expression profiles that support minimal residual disease monitoring (2).

Cytogenetic and molecular characterization of AML has become standard in clinical practice, as specific molecular alterations are now recognized as critical for therapeutic decision-making (3). *NPM1* mutation and *FLT3*-ITD are two typical examples of such biomarkers. Detecting *NPM1* mutation holds significant prognostic value, particularly in younger patients where it is associated with a more favorable clinical outcome (4). On the other hand, *FLT3*-ITD is associated with poorer prognosis but can be treated with specific inhibitors such as midostaurin (5) or quizartinib (6).

However, the advantages of molecular testing come with significant technical and financial challenges. Standard methods such as next generation sequencing (NGS) typically require expensive equipment, specialized personnel, and advanced data analysis pipelines (7). The reimbursement of this added financial burden is notoriously complex from the patient's perspective, raising critical equity concerns (8). Moreover, results are often only available weeks after the patient's initial management, rendering them ineffective for informing urgent treatment decisions (9).

Previous work found out that some cell expression profiles correlate with mutations of *NPM1* and *FLT3* (10–13), suggesting that prediction of these alterations could be achieved from flow cytometry data alone. As flow cytometry is already performed in routine care, such a prediction would come with no added costs and could be available in the very first days of patient's care.

Therefore, designing a reliable procedure to predict AML molecular profile from flow cytometry represents an opportunity of both better care and easier widespread adoption (Fig. 1).

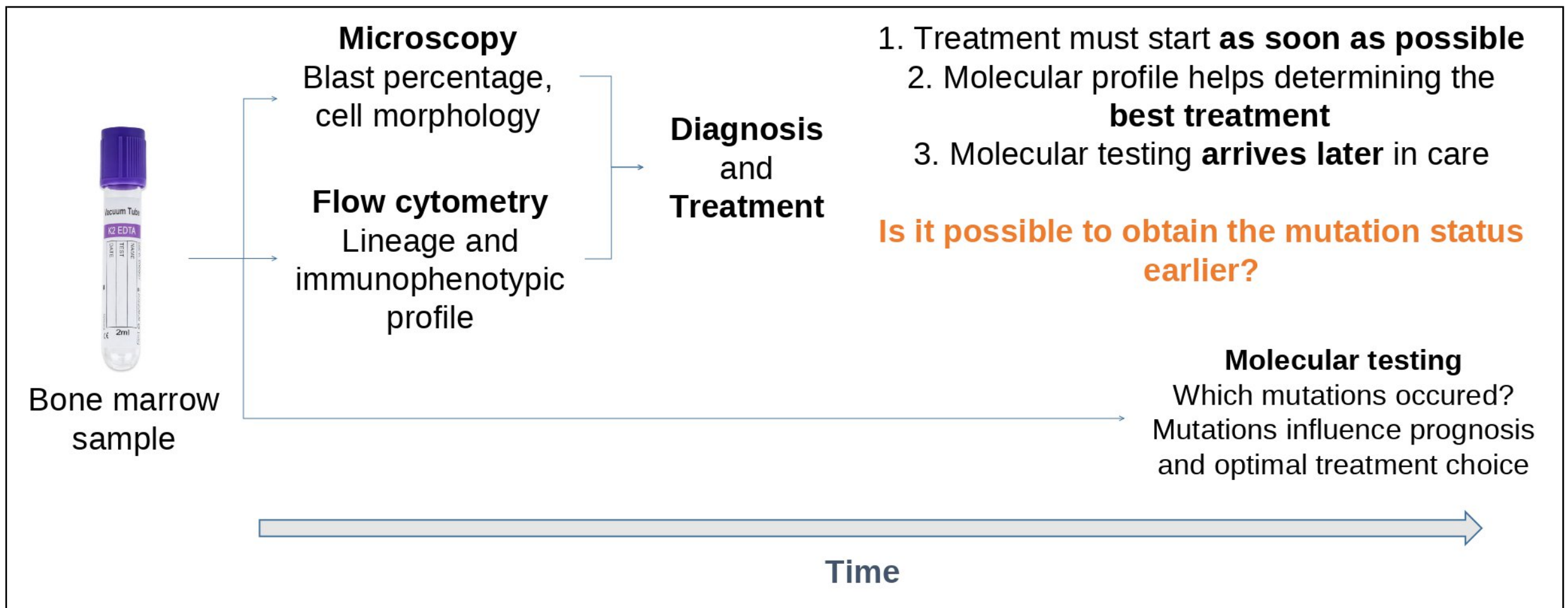

**Fig. 1. Diagnostic timeline in AML highlighting the delay between initial evaluation and molecular characterization.**

Microscopy and flow cytometry provide early information for diagnosis and treatment decisions, whereas molecular testing, although critical for identifying actionable mutations and guiding optimal therapy, is available later.

Molecular characterization from flow cytometry remains relatively underexplored. Recently, in (14), the authors reported the use of attention-based MIL models for mutation prediction in AML, achieving AUROCs of 0.80 for *NPM1* and 0.65 for *FLT3*-ITD in a cohort of 1 820 subjects. A detailed review of related work is provided in Supplementary Note S1.

The main contributions of this work are as follows:

- Demonstration that routine flow cytometry data analyzed using simple and interpretable machine learning models can reliably predict key molecular alterations in AML, specifically *NPM1* and *FLT3* mutation status, directly from standard clinical assays.
- Introduction of a decision tree-based MIL framework in which samples are modeled as bags of individual cells. Patient-level labels are inferred from cell-level predictions, as in single-instance learning. This pipeline was benchmarked against: (i) a random forest trained on readily available clinical variables; and (ii) a deep convolutional neural network adapted from (15) to handle multi-tube flow cytometry data. This comparison highlights trade-offs between model complexity, interpretability, and performance.

- Evaluation of the MIL framework using an independent dataset, providing an unbiased assessment of its generalizability and operational utility in real-world clinical settings.
- Demonstration that the MIL model's decision strategies are consistent with established hematological knowledge. Importantly, the framework identifies cell-level features driving predictions for *FLT3*-ITD and *NPM1* mutations, offering insights into their phenotypic signatures and potential therapeutic implications.

# Methods

## Patient cohorts

### Flow cytometry samples

#### *Exploration and validation datasets*

The exploration dataset comprised prospectively collected FCS files from 199 patients diagnosed with AML (excluding promyelocytic leukemia), included in ICAML study (NCT03449745) and treated at the Bordeaux University Hospital Center between July 01, 2019 and March, 31, 2020. For each patient, three distinct cytometry tubes, designated A, B, and C, were utilized. Tubes A and B assessed 12 cellular markers each, while tube C evaluated 10 markers (detailed in Table S2). Patients with missing tubes or missing mutation status were excluded from the study, yielding a dataset of 193 patients for *NPM1* and 197 for *FLT3*. An independent test cohort of 161 patients was also assembled from the DATAML Bordeaux database. Samples were collected between January 01, 2013 and December 31, 2018, with a set of tubes, antibodies and fluorochromes matching the exploration dataset. The study complied with the Declaration of Helsinki and DATAML is approved by French authorities [CNIL (N°915 285), CCTIRS (N°15.319)].

#### *Bone marrow sampling and preparation*

Bone marrow aspirates were collected in EDTA and processed within 24 hours of sampling. Samples underwent bulk red blood cell lysis using VersaLyse (Beckman Coulter), followed by antibody staining and a wash step. No intracellular staining was performed.

#### *Instrument configuration*

MFC was assessed on two Navios cytometers (Beckman Coulter), each equipped with 3 lasers (blue 488 nm, red 638 nm, violet 405 nm) and configured for 10-color, 12-detector acquisition with harmonized settings across both instruments. Instrument performance was verified daily using Flow-Check and Flow-Set fluorosphere beads and Immunotrol cells (all Beckman

Coulter). Compensation matrices were established manually in Kaluza 2.1 software (Beckman Coulter).

## Flow cytometry data pre-processing

Pre-processing of FCS files involved three sequential steps: (i) events at the signal detection margins were removed; (ii) fluorescence spillover was corrected using the compensation matrix embedded in each file; (iii) fluorescence intensities underwent $arcsinh(./5)$ transform. Pre-processed data were saved as new FCS files. Quality control plots for margin removal were generated for all samples, archived, and manually inspected to ensure data integrity.

For each patient, a random and uniform sample of $K$ cells was drawn from each tube. Two sampling sizes were used: $K$=100 and $K$=10 000, as detailed subsequently. Feature standardization was performed by computing the mean and standard deviation on the training data: within each cross-validation split of the exploration set during model comparison and on the full exploration set for the final model. Given input arrays of shape $(N, 3, K, M)$, where $N$ is the number of subjects, $3$ the number of tubes, $K$ the number of cells, and $M$ the number of cell features, means and standard deviations were computed per tube and marker by aggregating over the subject and cell dimensions, and used to standardize the corresponding left-out fold or test dataset.

## Clinical data pre-processing

Clinical variables were collected along with flow cytometry data: patient age, sex, and blast percentage (assessed via bone marrow smear microscopy). Categorical variables were one-hot encoded.

## Cohorts' description

Statistical comparisons between mutation groups were performed using the Kruskal-Wallis test for continuous variables and the Chi-squared test on contingency tables for categorical variables.

# Classification models for *FLT3*-ITD and *NPM1* mutation detection

## Proposed MIL model

A MIL approach for mutation detection was designed (Fig. 2): each patient is represented as a bag of cells, and the patient-level mutation status defines the bag label. A model is trained to predict the bag-level label from each cell individually. At inference time, bag-level predictions are obtained by aggregating cell-level predictions. This approach is usually referred to as single-instance learning in the MIL literature. In this study, decision trees were employed as classifiers due to their inherent interpretability.

### *Model training*

For each patient sample $X_n$ in the training set, cells were assigned the corresponding patient's mutation status ($y_n$). Cells from all training samples were then aggregated into a single dataset. A single, interpretable tree-based classifier was trained at the cell level for each individual tube.

### *Inference and score aggregation*

During inference, tubewise cell-level predictions were generated independently for each of the $K$ cells of an unseen sample. The resulting prediction scores were aggregated using mean pooling to derive a tube-level mutation score $\hat{z}_n$. Subsequently, scores from multiple tubes (3 for each patient) were combined via average pooling to produce a patient-level mutation score as a mutation probability. A binary mutation prediction $\hat{y}_n$ could be obtained by thresholding$\hat{z}_n$.

### *Integration of clinical data*

Optionally, tabular clinical data could be integrated in the prediction pipeline. Rather than directly pooling tube-level scores, these were fed along with the clinical features into an aggregator model, here a logistic regression. To prevent data leakage during training, the aggregator was trained on tube scores generated through an internal 5-fold cross-validation (stratified) scheme.

### *Cell-level interpretation*

For each tube-level prediction, a cell-level interpretation was generated. First, the decision paths associated with mutation detection were identified and characterized by the parameters of the trained decision tree. Individual cells were then projected onto the bi-plots corresponding to the path towards highest mutation probability, with their predicted mutation scores encoded as color gradient. This visualization approach facilitated the examination of cellular phenotypes that contribute the most to mutation detection.

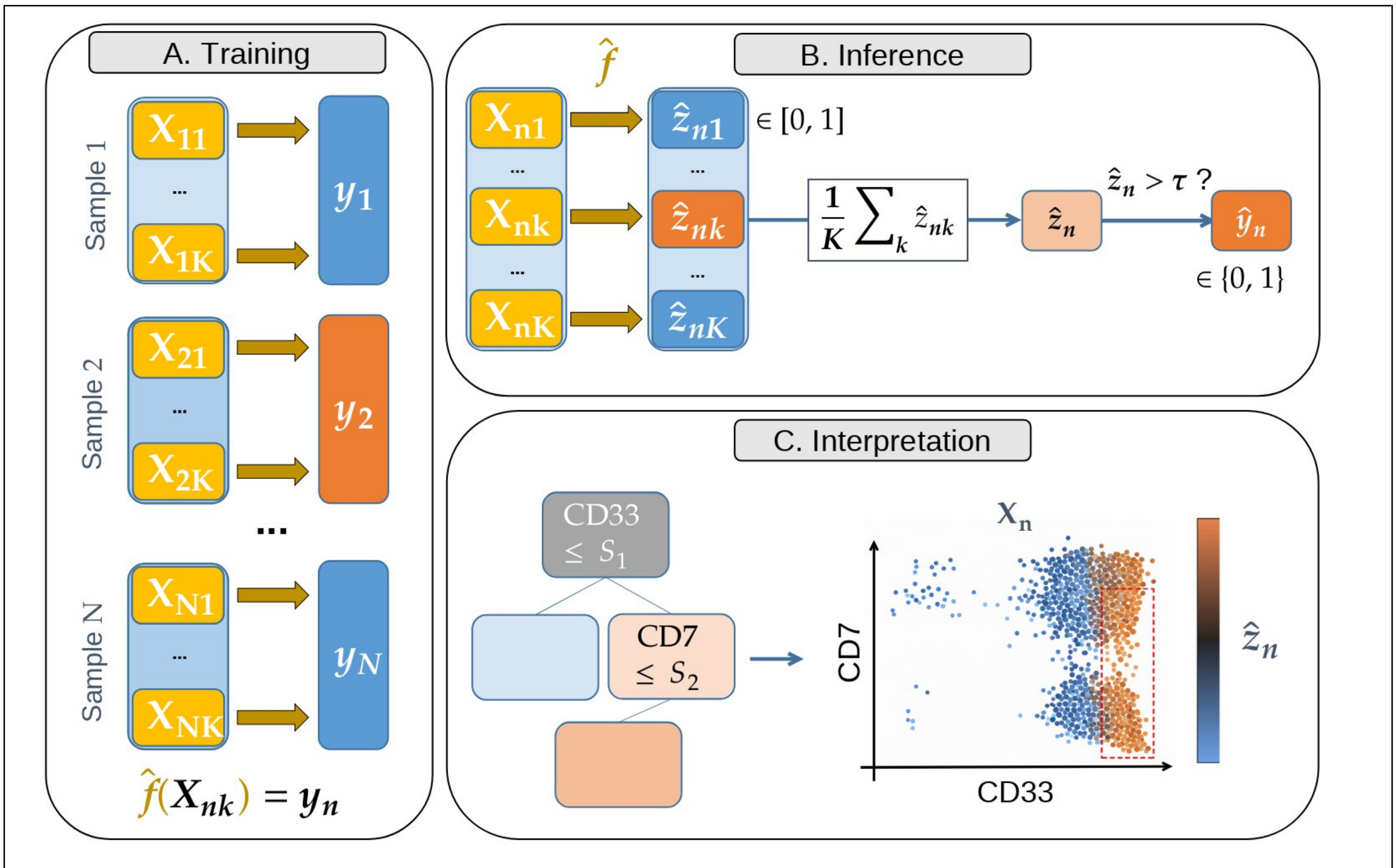

**Fig. 2. Multi-instance learning model for mutation detection**

A. Cell-level model training. For each sample $X_n$ derived from a single tube in the training dataset, cells were annotated according to the corresponding patient mutation status $y_n$. The annotated cells were then concatenated in a single array and used to train a cell-level classifier, here a decision tree.

B. Inference on unseen samples. To generate a prediction for an unseen sample, individual prediction scores of its K cells were computed. These scores were then averaged to produce a single sample-level prediction score $\hat{z}_n$. Binary predictions $\hat{y}_n$ were obtained from $\hat{z}_n$ by applying a decision threshold τ.

C. Cell-level interpretation. For any patient-level decision, an interpretability analysis was conducted: the decision path associated with the highest mutation probability was identified and characterized by the parameters of the trained decision tree, yielding relevant bi-plots for interpretation. Individual cell prediction scores were mapped to cell color intensity. The red rectangle indicates the gating regions defined by the decision tree, with boundaries determined by the learned split thresholds on the corresponding flow cytometry markers.

### *Hyper-parameter tuning on the exploration cohort*

A grid search with 5-fold cross-validation (stratified) was conducted to analyze the impact of the decision tree hyper-parameters and tube pooling function. To reduce computation time, the number of input cells $K$ was limited to 100 cells per tube. The search space included the following parameters: maximum tree depth (3, 4), minimum samples per leaf (1, 100, 500), cost-complexity pruning parameter alpha (logarithmically spaced between 1 and $10^{-6}$), and tube pooling functions (maximum, mean, minimum). Models were evaluated on a per-fold basis, and the best-performing hyper-parameter configuration was selected based on mean AUROC.

The classification decision threshold $\tau$ was then tuned to optimize the precision-recall trade-off. For that purpose, 100 candidate thresholds were evaluated via 5-fold cross-validation (stratified) using the $F_\beta$ score with $\beta = 0.5$, which assigns twice the weight to precision relative to recall.

## Alternative approaches

The MIL approach was evaluated against two alternative strategies representing different levels of model complexity: a baseline model trained on routinely available clinical variables and a deep CNN derived from the computational cytometry literature.

### *Baseline model using tabular clinical data*

The baseline model was a random forest classifier trained on the above-mentioned clinical data.

### *Deep CNN for flow cytometry data*

The deep CNN architecture proposed in (15) was adapted to accommodate a multi-tubes setting using depth-wise separable convolutions (Fig. 3). The original optimal hyper-parameters were retained, with a fixed input size of 10 000 cells.

In experiments combining cytometry and clinical data, a patient-specific clinical feature vector was concatenated with the convolution-derived embedding, and fed into the fully connected layers.

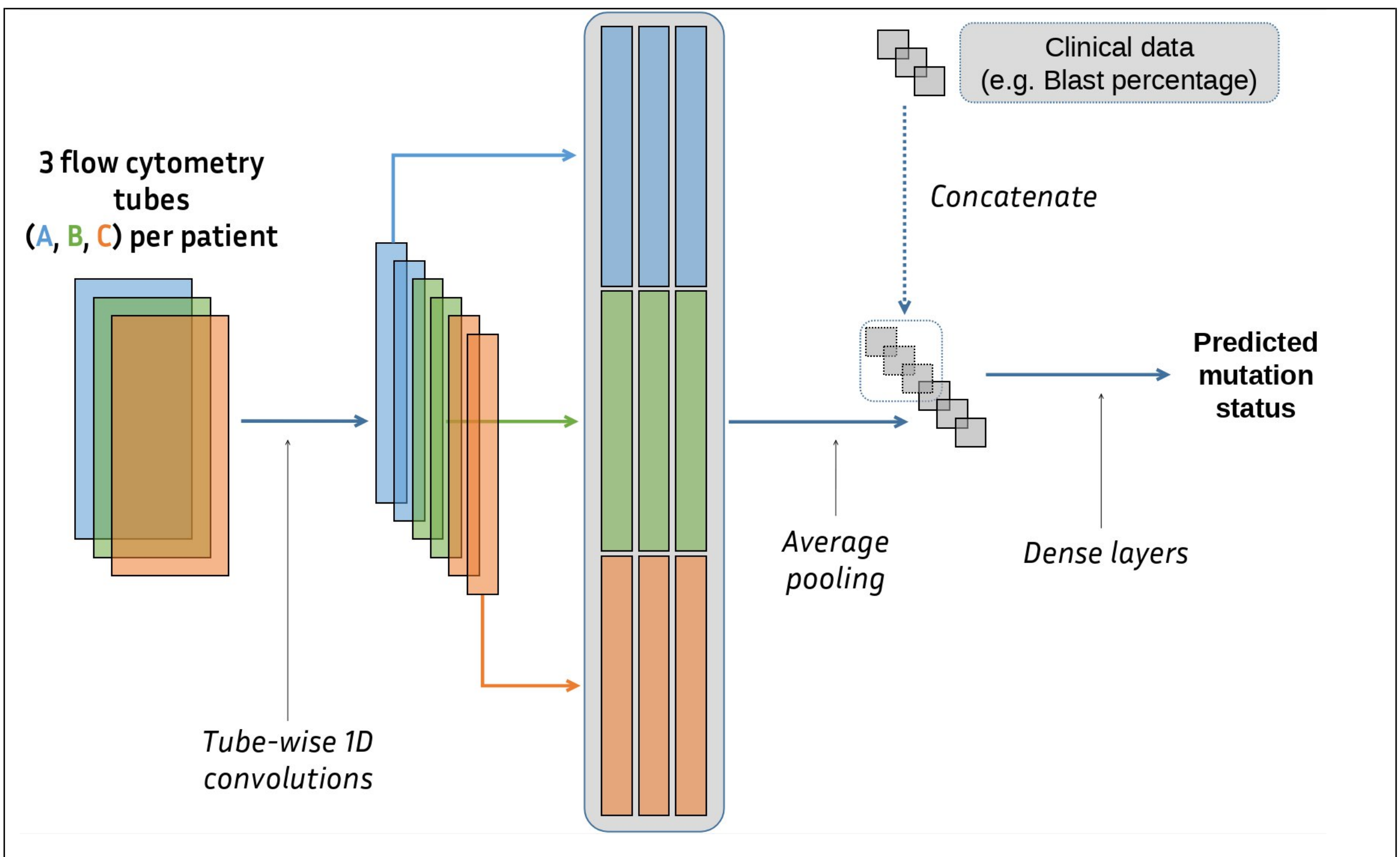

**Fig. 3. Deep CNN architecture for flow cytometry analysis**

The proposed architecture is adapted from the work of (15) and designed to process data from multi-tube flow cytometry. The model accepts input samples structured as tensors of shape (C, K, D), where C denotes the number of tubes, K the number of cells per tube and D the number of features per cell. Each cell was independently processed using tube-wise 1D convolutional layers. The resulting cell-level representations were then aggregated via an average pooling layer, which transforms the C x K cell vectors into a single patient-level embedding. Optionally, this vector can be concatenated with tabular clinical data (such as age or blast percentage). Fully connected layers with C nodes generate the final decision.

## Experimental evaluation protocol

Model development and evaluation followed a two-step procedure. First, all candidate models were compared using default parameter settings on the exploration dataset in order to assess their relative performance under comparable conditions. Then, the proposed MIL model was further optimized and its generalization performance was evaluated on the independent external test set.

### Training and inference via cross-validation

All models were trained and validated using a 10-fold cross validation scheme (stratified). For the CNN and the baseline model, training sample importance was weighed by the inverse of class frequencies to reduce class imbalance. Cytometry-based models were fed $K = 10\ 000$ cells per sample to allow a fair comparison across approaches.

The random forest was trained using scikit-learn's default hyper-parameters. The CNN was configured according to recommendations outlined in the original study (15). For the proposed MIL model, mean aggregation was applied at both the cell and tube levels, and model complexity was constrained by setting the *min_samples_leaf* parameter to 1 000.

Cross-validation performance results are reported as the mean AUROC along with the standard deviation (SD) across folds.

### Statistical comparison of classifiers

To statistically compare classifiers performance without assuming normality or independence between folds (conditions not inherently guaranteed in cross-validation settings), pairwise permutation tests were conducted (16), as detailed in Supplementary Note S3.

## Inference on external test set

The MIL model with optimized hyper-parameters was evaluated on the independent test set to assess its generalization performance. To estimate variability in performance, we evaluated the model using repeated random subsampling.

AUROC, Positive Predictive Value (PPV), and Negative Predictive Value (NPV) were calculated by repeating the inference and evaluation procedures across 100 independent trials, each initialized with a distinct random seed for cell sampling. The mean and standard deviation (SD) of each metric, aggregated over all trials, were reported. This evaluation was also conducted independently for each tube (A, B and C) to determine which tubes retained discriminative power on external data and whether the biological patterns learned during training effectively transferred to the independent cohort.

# Hardware and implementation

Flow cytometry data were preprocessed in R (v. 4.5.1) using the flowCore v. 2.20.0 (ref. 17) and PeacoQC v.1.18.0 (ref. 18) packages. Machine learning models were implemented in Python 3.12.2 with PyTorch (v. 2.7.0) for the deep CNN and scikit-learn v.1.6.1 (ref. 19) for random forest classification, cross-validation and threshold optimization. Training of the deep CNN was performed on a local GPU hardware. Other computations were performed on local community hardware. All the scripts used for this analysis are available in a public repository[1].

[1] https://github.com/jonathan-legrand/cytometry-molecular-characterization

# Results

## Description of the exploration cohort

Patients' characteristics are given in Table 1 for the exploration cohort and in Table S4 for the test cohort.

| | **Grouped by *NPM1* status** | | | |
|---|---|---|---|---|
| | **Overall** | ***NPM1*mut** | ***NPM1*wt** | **P-Value** |
| **n** | 193 | 55 | 138 | |
| **Age, median [Q1,Q3]** | 63 [51,69] | 64 [52,69] | 62 [51,69] | 0.825 |
| **Blasts bone marrow (%), median [Q1,Q3]** | 63 [39,83] | 77 [55,93] | 56 [35,80] | <0.001 |
| **Number of events tube A, median [min,max]** | 61474 [21612,920992] | 49670 [21612,920992] | 65944 [21790,655085] | 0.015 |
| **Number of events tube B, median [min,max]** | 57930 [17218,929315] | 46600 [17218,929315] | 66214 [21102,629922] | 0.002 |
| **Number of events tube C, median [min,max]** | 56372 [21131,904603] | 41107 [21131,904603] | 65247 [22349,549173] | 0.001 |
| **Female, n (%)** | 88 (100) | 28 (32) | 60 (68) | 0.438 |
| **Male, n (%)** | 105 (100) | 27 (26) | 78 (74) | |
| ***FLT3*-ITD, n (%)** | 37 (100) | 19 (51) | 18 (49) | 0.001 |
| ***FLT3* No ITD, n (%)** | 156 (100) | 36 (23) | 120 (77) | |

**Table 1: Demographic and clinical features of the exploration cohort**
Statistical comparisons between groups were performed using the Kruskal-Wallis test for continuous variables and the Chi-squared test on corresponding contingency tables for categorical variables.

## Comparison of classifiers performances

The Figure 4 illustrates the classification performance of the models on *NPM1* and *FLT3*-ITD mutation detection.

### *NPM1* mutation detection

For *NPM1* mutation prediction task (Fig. 4A), the baseline model achieved a mean AUROC of 0.62 across cross-validation folds (SD = 0.10). In contrast, both the CNN with clinical data and the cytometry-only CNN demonstrated improved and comparable performance, with mean AUROCs of 0.95 (SD = 0.04) and 0.94 (SD = 0.07), respectively. The MIL models achieved similarly high performance, with a mean AUROC of 0.96 (SD = 0.05) for the cytometry-only variant and 0.95 (SD = 0.06) for the model incorporating clinical variables.

All cytometry-based classifiers significantly outperformed the baseline ($P < 0.05$, ΔAUROC > 0.33). However, no statistically significant differences in pairwise comparisons were observed among the cytometry-based classifiers.

### *FLT3*-ITD mutation detection

For *FLT3*-ITD detection task (Fig. 4B), the baseline model achieved a mean AUROC of 0.70 across cross-validation folds (SD = 0.10). The cytometry-only CNN yielded an improved mean AUROC of 0.80 (SD = 0.12); however, this improvement did not reach statistical significance ($P$=0.06). Incorporation of clinical variables into the CNN resulted in comparable performance (mean AUROC 0.80, SD = 0.12). The MIL models achieved the highest performance, with mean AUROCs of 0.86 (SD = 0.10) using cytometry data alone and 0.87 (SD = 0.10) when combining cytometry and clinical variables.

With the exception of the cytometry-only CNN, all cytometry-based classifiers significantly outperformed the baseline (P<0.05, ΔAUROC > 0.14). However, no statistically significant differences in pairwise comparisons were observed among the cytometry-based classifiers.

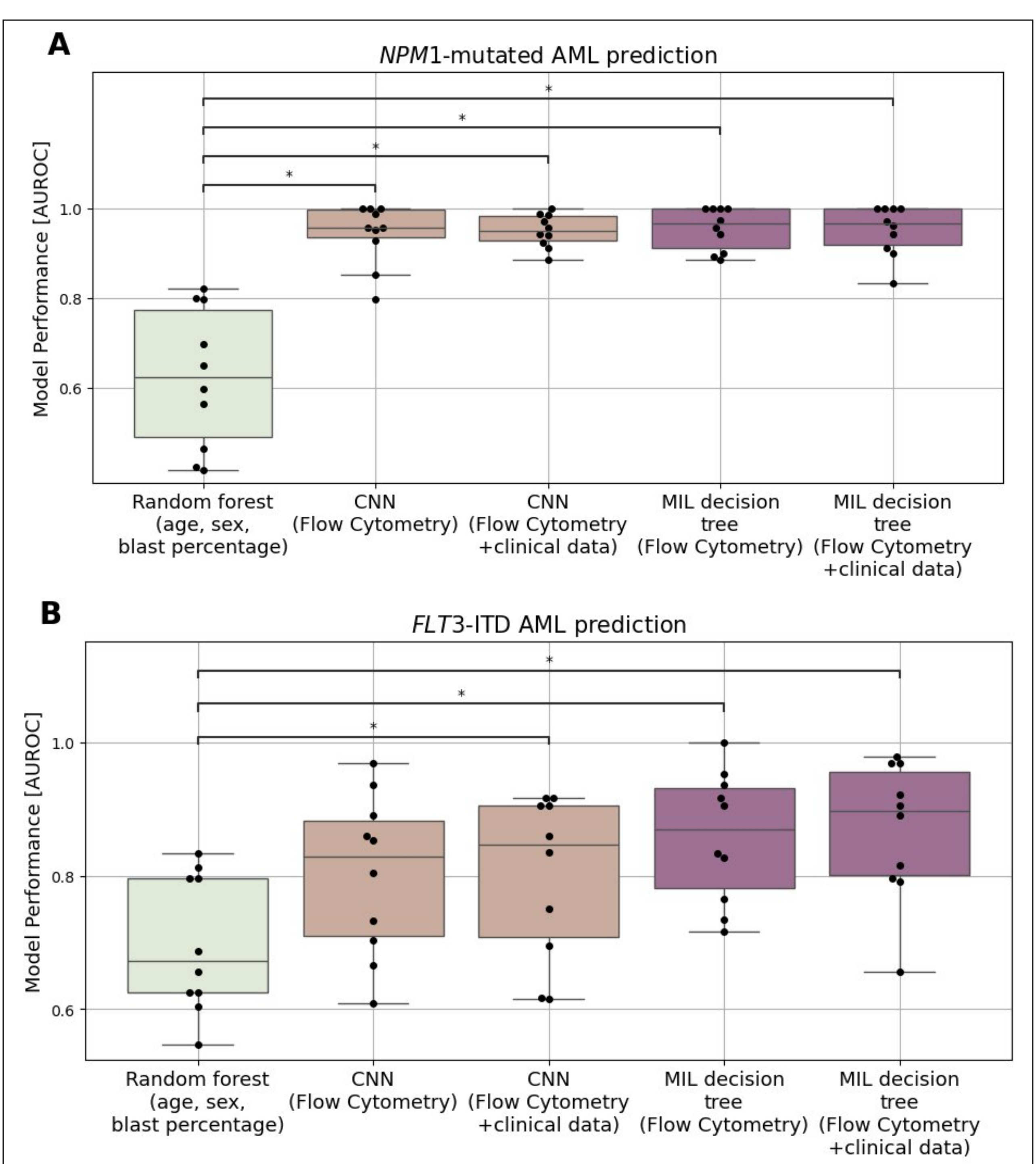

**Fig. 4. Models' performance on left-out validation folds**

All classifiers were trained and tested within the same cross-validation framework. Performance distributions are visualized using box plots, where the box represents the median and interquartile range (IQR), and the whiskers extend to show to the minimum and maximum values within 1.5xIQR. Statistical significance was assessed using a permutation test with 1 000 shuffles. *P < 0.05

(A) *NPM1* classification results. All flow cytometry-based models significantly outperformed the

random forest baseline. No significant pairwise difference was observed among cytometry-based classifiers.

(B) *FLT3*-ITD classification results. With the exception of the cytometry-only CNN, all cytometry-based classifiers demonstrated significantly superior performance compared to the random forest baseline. No significant pairwise difference was observed among cytometry-based classifiers.

## Hyper-parameter tuning and model performance

The set of best hyper-parameters for each task is available in Table S5. Performance metrics obtained on the external test set are summarized in Table 2.

| Mutation | Method | Mean performance [SD] | | |
|---|---|---|---|---|
| | | AUROC | PPV | NPV |
| *NPM1*-mut | Tube A | **0.90 [0.0]** | 0.83 [0.02] | 0.86 [0.01] |
| | Tube B | 0.89 [0.0] | 0.83 [0.02] | **0.87 [0.01]** |
| | Tube C | 0.88 [0.0] | 0.79 [0.03] | 0.85 [0.01] |
| | All tubes | **0.90 [0.0]** | **0.87 [0.02]** | 0.81 [0.01] |
| *FLT3*-ITD | Tube A | **0.82 [0.01]** | **0.68 [0.06]** | 0.86 [0.01] |
| | Tube B | 0.77 [0.01] | 0.57 [0.03] | 0.91 [0.01] |
| | Tube C | 0.81 [0.01] | 0.67 [0.06] | 0.87 [0.01] |
| | All tubes | 0.8 [0.01] | 0.55 [0.03] | **0.92 [0.01]** |

**Table 2: Classifier performance on the external test set**

Mean and SD estimates were obtained by repeating the inference and evaluation procedures 100 times with different seeds for cell sampling.

## Cell-level interpretation on the test dataset

Figure 5 presents the decision tree for detecting *NPM1* mutation in tube A (up), together with the cell-level interpretation bi-plots generated for each tube (down). All tube-wise decision trees are displayed in Figure S6. Across all tubes, a “CD33 high/CD34 low” criterion consistently characterized the main cell population associated with *NPM1* mutation detection (Fig.5, up; Fig.

S1). Tube-specific patterns then induced further stratification of this population in terms of mutation probability. In tube A, “CD14 low/CD65 low” characterized the largest pro-mutation subpopulation also having the highest mutation probability (Fig.5, down-left). Additional discriminative positive paths included “CD33 high/CD34 low/CD14 high/CD45 high”. In tube B, “CD64-/CD38+” led to the highest mutation probability (Fig.5, down-center). Moreover, the tree shows that a three-fold splitting based only on CD64 was essentially enough to stratify cells into pro-mutation subpopulations, intermediate CD64 corresponding to the largest subpopulation having a lower probability. In tube C, all “CD33 high/CD34 low” were pro-mutation, and “CD36 high/CD2 low” was associated with the subpopulation with the highest mutation probability (Fig.5, down-right), although involving fewer cells. The tree indicates that a three-fold splitting based only on CD36 already allows to stratify this population in terms of mutation probability.

Figure 6 presents the decision tree used for detecting *FLT3*-ITD mutation detection in tube A (up), together with the cell-level interpretation bi-plots generated for each tube (down). The tube-wise 3-steps decision trees are displayed in Figure S7. Across all tubes, the most consistent pattern found in pro-mutation cells was “CD33 high/SS-” (Fig. 6, bottom panel, up-left and up-right). Additional tube-specific phenotypic patterns were identified: in tubes A and C, cells with the highest *FLT3*-ITD probability exhibited low CD34 expression (Fig. 6, down-right); In tube B, *FLT3*-ITD-positive cells were further stratified by CD123, with the highest probability for CD123 high/CD56 low cells (Fig. 6, down-middle).

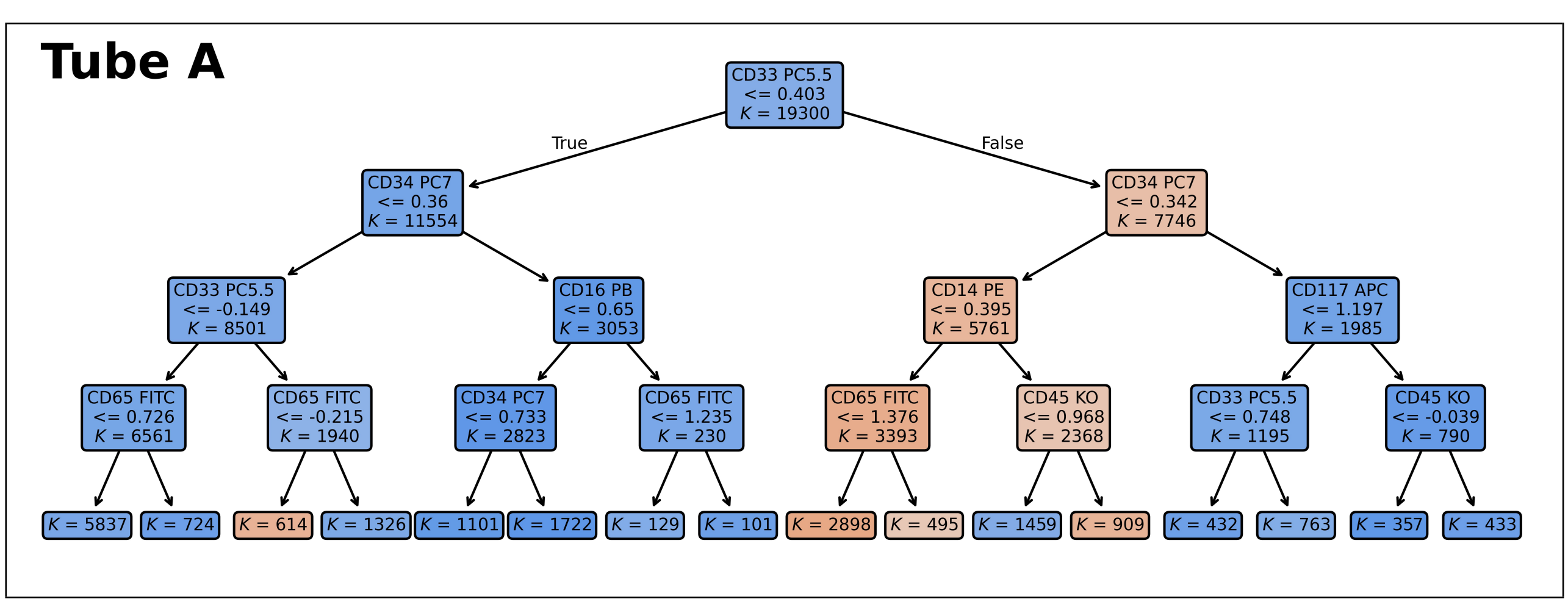

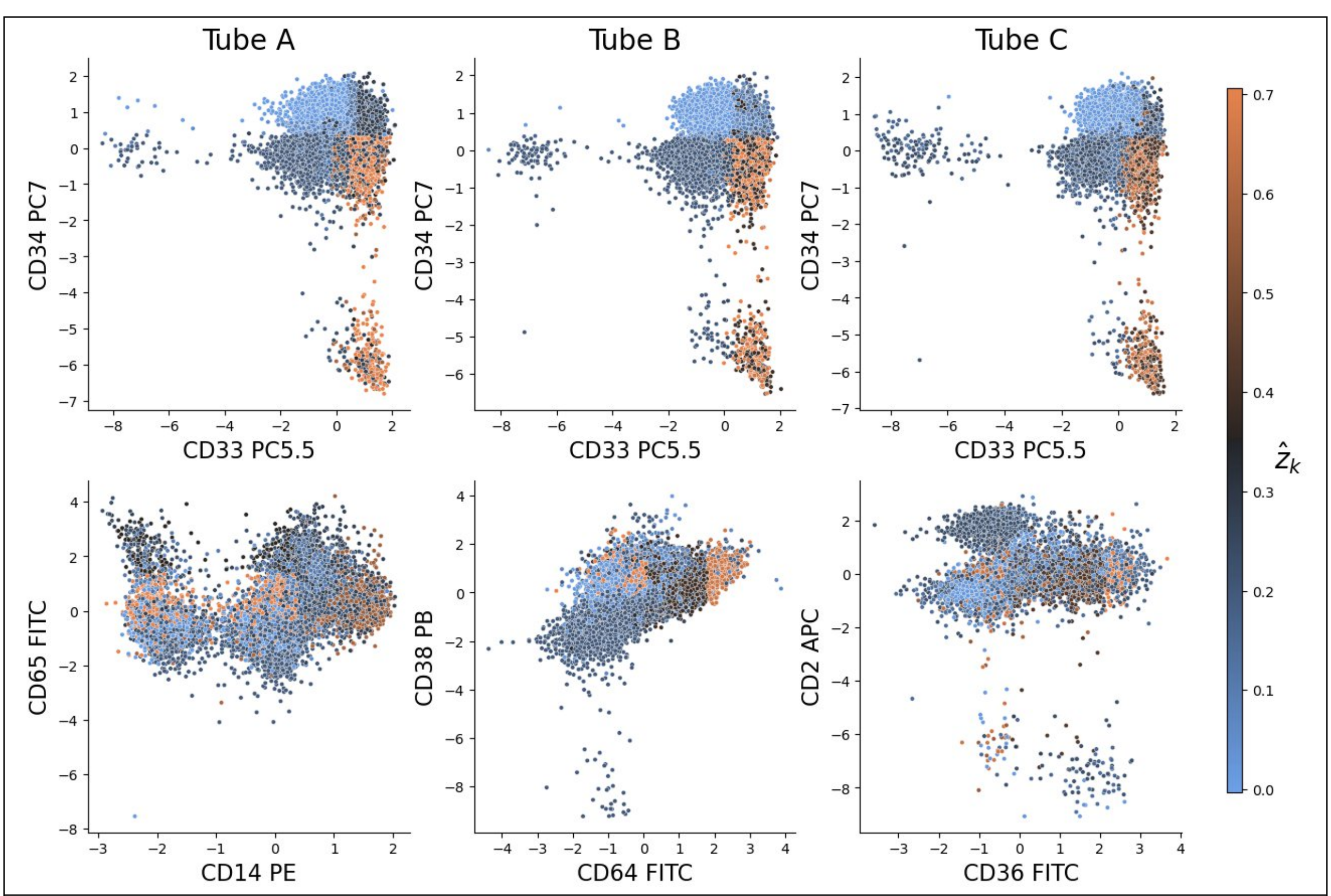

**Fig. 5. Cell-level interpretation for *NPM1* mutation detection on the test dataset**

Up: Decision tree obtained for Tube A. Color indicates the majority class at each node (orange: mutation-positive; blue: mutation-negative). K denotes the number of training cells at a node prior to splitting. During inference, probability estimates are calculated as the proportion of mutation-positive training cells within the corresponding leaf node. Thresholds are reported as standardized dimensionless arcsinh-transformed marker intensities. Down: Each column shows a tube-specific cell-level interpretation derived from a 4-steps decision tree trained to predict *NPM1* mutation status. Cell color encodes the predicted mutation probability for each cell ($\hat{z}_{nk}$), centered around the tuned decision threshold ($\tau = 0.35$). Axes show standardized, dimensionless arcsinh-transformed marker intensities.

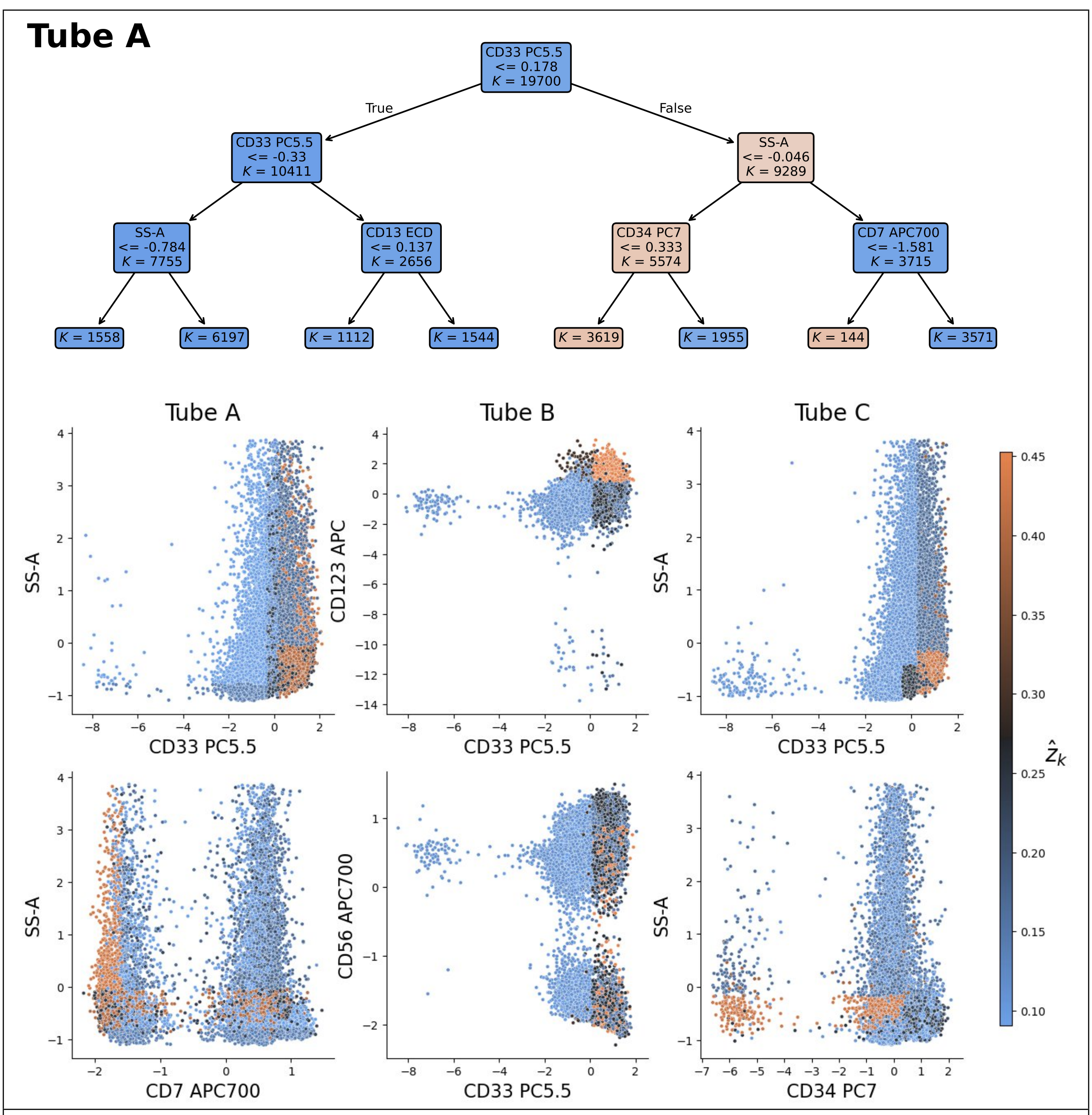

**Fig. 6. Cell-level interpretation for *FLT3*-ITD detection on the test dataset**

Up: Decision tree obtained for Tube A. Color indicates the majority class at each node (orange: mutation-positive; blue: mutation-negative). K denotes the number of training cells at a node prior to splitting. During inference, probability estimates are calculated as the proportion of mutation-positive training cells within the corresponding leaf node. Thresholds are reported as standardized dimensionless arcsinh-transformed marker intensities. Down: Each column shows a tube-specific cell-level interpretation derived from a 3-steps decision tree trained to predict *FLT3*-ITD mutation status. Cell color encodes the predicted mutation probability for each cell ( $\hat{z}_{nk}$), centered around the optimized decision threshold ($\tau = 0.27$). Axes show standardized, dimensionless arcsinh-transformed marker intensities.

# Discussion

Cytogenetic and molecular characterization plays a central role in the clinical management of AML. Alterations such as *NPM1* mutation and *FLT3*-ITD directly inform risk stratification and guide important therapeutic decisions. In middle-aged patients, who represent the peak incidence of the disease, therapeutic decisions sometimes need to be made urgently, before prognostic factors are available. Predicting mutational status is crucial in this context.

For instance, *NPM1* patients are often leukocytotic and may require emergency treatment within the first hours after diagnosis, before their molecular profile is known. Conversely, other patients with a similar clinical presentation may have molecular features associated with high resistance to intensive chemotherapy and could benefit more from targeted therapies based on venetoclax. The strategy developed in this study could help improve the management of these situations.

In addition, the introduction of theranostic targeted therapies which offer new therapeutic opportunities, such as *IDH1* or *FLT3* inhibitors, is supported by biological tests that are sometimes expensive and not always readily accessible. In this context, the ability to infer molecular status directly from flow cytometry data would represent a valuable complementary approach to NGS, enabling faster and more accessible characterization.

However, predicting molecular alterations from flow cytometry data remains challenging. The datasets analyzed in this study were generated in routine clinical practice over several years, thereby reflecting both technical variability and biological heterogeneity. Moreover, immunophenotypic patterns in AML are continuous, multidimensional, and shaped by substantial clonal diversity, making robust genotype–phenotype associations difficult to capture.

## Comparison of classifiers and methodological considerations

Despite the above-mentioned challenges, both implemented cytometry-based classifiers achieved strong predictive performance, demonstrating that clinically relevant molecular information can be recovered from routine flow cytometry data alone. The performance of the proposed tree-based MIL classifier on the test set is on par with previously reported results for image-based mutation detection: a 0.92 test AUROC was reached in (20) for *NPM1*, while (21) reported 0.86 AUROC for *NPM1* and 0.79 for *FLT3*-ITD. In the cytometry literature, a lower performance was reported in (14) in a cross-validation setting for both tasks (0.80 for *NPM1* mutation, 0.65 for *FLT3*-ITD), although rigorous comparison of methods is not feasible when evaluating on different datasets.

Both the proposed tree-based MIL model and the existing deep CNN implement a conceptually similar strategy: cell-level classification followed by a pooling step at the sample level. Notably, despite its simplicity, the tree-based approach demonstrated performance comparable to that of the deep CNN for the two tasks implemented in this study. While the generalizability of this finding to other mutation detection tasks remains to be assessed, it is worth noting that tree-based models have been shown to outperform neural networks in small data regimes, particularly for tabular data (22).

Importantly, adding clinical metadata such as the percentage of blasts in bone marrow as evaluated on bone marrow smear images did not significantly improve the performance of the models in any of the prediction tasks. This may in fact be advantageous, as incorporating myelogram results would introduce additional delays and complexity into the workflow, making the pipeline less efficient from an operational standpoint.

## Interpretability and biological plausibility of tree-based MIL models

Interpreting the prediction pipeline provides insight into the contribution of individual markers to mutation prediction. With the proposed tree-based model, interpretation is done by examining the explicit decision rules learned by the model, which directly inform the probability of the mutation being present. It is important to note that the approach relies on a specific realization of cell sampling, which serve as input for the classification algorithm. The classification tree constructed for a given tube is thus anticipated to exhibit variability across successive realizations, with each realization potentially introducing novel and pertinent information.

For *NPM1*-mutated AML, the MIL model consistently identified a CD33 high/CD34 low profile across tubes, with tube-specific refinements primarily involving CD14, CD64 and CD36, followed by secondary contributions from CD65, CD38, and CD2, respectively. These findings align with previously reported immunophenotypes. Importantly, beyond identifying individual marker associations, the tree structure naturally captures phenotypic heterogeneity within a single mutation class. In tube A, the model stratifies the CD33$^{+}$/CD34$^{-}$ population into two biologically distinct subgroups: a monocytic branch (CD14$^{+}$/CD45 bright) and a non-monocytic branch (CD14$^{-}$/CD65$^{-}$), consistent with the observation that *NPM1*-mutated AML encompasses both monocytic and myeloid differentiation subtypes (10,13). The CD34$^{-}$/CD14$^{-}$ profile of the latter branch is also concordant with the findings of (23).

Beyond validating known patterns, the model also facilitates the identification of previously unreported associations, including a potential link between *NPM1* mutations and CD36 expression. One potential explanation is that *NPM1*-mutated leukemias frequently exhibit monocytic subtype (FAB M4 and M5) (ref. 13), and monocytes commonly express CD36 (ref. 24). This suggests that the model may be capturing features related to monocytic differentiation.

In *FLT3*-ITD AML, the MIL model identified cell populations characterized by high CD33 expression and low side scatter, with variable contributions from CD34 and CD123 depending on the tube. These results are consistent with prior reports of increased CD33 and CD123 expression in *FLT3*-ITD AML (11). Notably, the model identified CD34$^{-}$ as a defining marker for *FLT3*-ITD, likely reflecting the co-occurrence of *FLT3*-ITD with *NPM1* mutations and their combined impact on immunophenotype (10). To further investigate this association, the model was retrained to predict *FLT3*-ITD on a dataset that excluded *NPM1*-mut patients. In this setting, CD34- was no longer associated with *FLT3*-ITD, supporting the hypothesis that the observed relationship is driven by mutation co-occurrence rather than a direct effect of *FLT3*-ITD alone. Details of this analysis, including scores on the test dataset and model interpretation, can be found in  Supplementary information (respectively Table S8 and Figure S9).

While deep CNNs can in principle be interpreted, the method proposed in (15) relies on a post-hoc procedure. This approach requires iterative up-sampling of each cell in the dataset and performing a forward pass at every step, resulting in substantial computational overhead. Furthermore, its applicability and reliability in the context of hematological malignancies remain uncertain (25). In contrast, the interpretation of the proposed tree-based classifier is immediate, as it is inherently linked to the explicit decision rules learned by the model.

# Towards prediction in operational setting

## Performance and PPV

Beyond global performance metrics, mutation-specific predictive values must be considered to assess the translational potential of this approach. From an operational standpoint and for these two specific mutations, the PPV is particularly important, as it directly influences clinical decision-making. Consequently, the classifier’s decision threshold was optimized using the $F_\beta$ score with $\beta = 0.5$, which assign twice the weight to precision compared to recall. This metric ensures that the classifier operates at a decision boundary where false positives are penalized more heavily given their clinical implications.

However, in practice, the PPV for *FLT3*-ITD detection remained moderate (up to 0.68 using tube A) which currently limits the use of the model as a stand-alone confirmatory test. In contrast, the PPV for *NPM1* characterization reached 0.87 (all tubes combined), which may be attributed to the relatively homogeneous phenotype associated with *NPM1* mutations.

On the other hand, across both tasks, the model achieved strong performance in identifying non-mutated patients, reaching a NPV of 0.87 for *NPM1* (tube B) and 0.92 for *FLT3* (all tubes combined). These results suggest a potential clinical utility for flow cytometry-based models as rapid rule-out diagnostic tools. This ability to confidently exclude mutations is particularly important for some other alterations associated with poor prognosis, such as secondary-type or *TP53* alterations, when determining whether intensive chemotherapy is appropriate or not.

## Designing evidence-based panels for molecular characterization

An important practical implication of these findings concerns panel design. The flow cytometry tubes used in this study were optimized for measurable residual disease monitoring, as no panel dedicated to molecular characterization is currently available in clinical practice. The three-tube protocol enabled evaluation of 24 markers, but inspection of the learned decision trees showed that only a limited subset consistently contributed to classification. CD33, CD34, CD14, CD65, CD38, CD2, CD36, CD64, CD45, CD123 and SS were sufficient to identify cell populations associated with the highest mutation probabilities for both *NPM1* and *FLT3*. These findings suggest that a dedicated, more compact panel could achieve similar predictive performance while simplifying workflows, reducing costs, and facilitating broader clinical implementation.

## Model generalization across cohorts

It is important to emphasize that the exploration and test cohorts used in this study exhibit substantial differences. The exploration cohort was collected within the framework of a research study (NCT03449745), which facilitated enhanced standardization and the acquisition of a higher number of cells per sample. In contrast, the test cohort reflects fully routine clinical practice, characterized by inherent variability and operational constraints. Despite these differences, the proposed tree-based MIL model successfully generalized to the testing cohort.

It must be acknowledged, however, that transferring models across laboratories remains a significant challenge due to differences in fluorochromes and sample preparation protocols. Future work should assess performance degradation when training and testing across distinct protocols and explore domain adaptation strategies for supervised learning in the context of flow cytometry.

The predictive framework presented here was applied to conventional flow cytometry panels, comprising 9 to 10 surface markers per tube. The recent development of spectral flow cytometry, which enables simultaneous acquisition of 28 or more parameters per cell (26), offers a natural extension of this approach. A higher-dimensional phenotypic space would provide finer resolution of cell subpopulations and could improve prediction accuracy, particularly for mutations where current performance remains moderate, such as *FLT3*-ITD, or for mutations not yet explored, such as *TP53* or *IDH1/2*.

In addition, all markers used in this study were restricted to cell surface antigens, as no intracellular staining was performed. Incorporating intracellular markers such as phosphorylated signaling intermediates downstream of mutated kinases, or metabolic markers reflecting mutation-driven reprogramming could capture biological features that are not accessible through surface phenotyping alone.

Finally, this analysis revealed that the tree-based model can identify phenotypically distinct subpopulations within a single mutation class, as illustrated by the monocytic and non-monocytic branches among *NPM1*-predicted cells. While consistent with the known co-occurrence of mutations and their combined effects on immunophenotype, this observation raises the possibility that cell-level prediction models trained on sufficiently large and well-annotated datasets could eventually resolve distinct mutational subclones within individual patient samples, motivating further research in multicentric cohorts and single-cell genotyping approaches.

Overall, these results demonstrate the potential of interpretable machine learning to provide early molecular insights in AML, enabling more timely and accessible precision medicine. A natural extension of this work is the application of the proposed approach to other clinically relevant genes, such as *TP53* and *IDH1/2*. The increasing availability of higher-dimensional spectral cytometry data may also offer opportunities to enhance predictive performance. Finally, the development of robust domain adaptation strategies will be essential for transferring models across laboratories and protocols, ultimately supporting their widespread adoption and clinical impact.

# Acknowledgements

The CytoFLAM project has received the joint financial support of the Inria Centre at the University of Bordeaux and of the Nouvelle-Aquitaine Region in the scope of their “Research and Platform” call. This project was partially funded by the AOI of the Bordeaux University Hospital. We would like to thank the clinical data management unit of Toulouse University Hospital.

During the preparation of this manuscript, the authors used Le Chat by Mistral AI, Claude by Anthropic and ChatGPT by OpenAI to refine the clarity and language of the text. After using this tool, the authors carefully reviewed and edited the content and take full responsibility for the final version.

# Authors contribution

CE was responsible for conceptualization of the study, funding acquisition, supervision, reviewing and editing the draft. PYD was responsible for conceptualisation of the study, reviewing and editing the draft. AB conducted the data curation. BDS was responsible for conceptualization of the study, funding acquisition, supervision, reviewing and editing the draft. AM contributed to conceptualization of the study, data curation, investigation, reviewing and editing the draft. JL conducted the development of methodology and software, performed the formal analysis and wrote the original draft.

# Conflict of interest

The authors declare no competing financial interests in relation to this work.

# Data availability statement

The datasets analysed during the current study are not publicly available due to privacy or ethical restrictions but are available from the corresponding author on reasonable request.

## Ethics approval and consent to participate

This study was conducted in accordance with the Declaration of Helsinki. The exploration cohort was collected as part of the ICAML study (ClinicalTrials.gov identifier NCT03449745). The validation cohort (DATAML) was approved by the relevant French regulatory authorities: the Commission Nationale de l'Informatique et des Libertés (CNIL, reference N°915 285) and the Comité Consultatif sur le Traitement de l'Information en matière de Recherche dans le domaine de la Santé (CCTIRS, reference N°15.319). Informed consent to participate was obtained from all patients included in the study.

## Consent for publication

Not applicable.

# References

1. Döhner H, Weisdorf DJ, Bloomfield CD. Acute Myeloid Leukemia. N Engl J Med. 2015;373(12):1136–52. doi:10.1056/NEJMra1406184

2. Döhner H, Wei AH, Appelbaum FR, Craddock C, DiNardo CD, Dombret H, et al. Diagnosis and management of AML in adults: 2022 recommendations from an international expert panel on behalf of the ELN. Blood. 2022;140(12):1345–77. doi:10.1182/blood.2022016867

3. Döhner H, DiNardo CD, Appelbaum FR, Craddock C, Dombret H, Ebert BL, et al. Genetic risk classification for adults with AML receiving less-intensive therapies: the 2024 ELN recommendations. Blood. 2024;144(21):2169–73. doi:10.1182/blood.2024025409

4. Liu VM, Othus M, Ries RE, Naru J, Pogosova-Agadjanyan EL, Appelbaum FR, et al. Evaluation of ELN2022 Risk Stratification in NPM1 Mutated AML: A Study from the Fred Hutch and SWOG. Blood. 2024;144(Supplement 1):4309. doi:10.1182/blood-2024-198506

5. Stone RM, Mandrekar SJ, Sanford BL, Laumann K, Geyer S, Bloomfield CD, et al. Midostaurin plus Chemotherapy for Acute Myeloid Leukemia with a FLT3 Mutation. N Engl J Med. 2017;377(5):454–64. doi:10.1056/NEJMoa1614359

6. Erba HP, Montesinos P, Kim HJ, Patkowska E, Vrhovac R, Žák P, et al. Quizartinib plus chemotherapy in newly diagnosed patients with FLT3-internal-tandem-duplication-positive acute myeloid leukaemia (QuANTUM-First): a randomised, double-blind, placebo-controlled, phase 3 trial. The Lancet. 2023;401(10388):1571–83. doi:10.1016/S0140-6736(23)00464-6

7. Bacher U, Shumilov E, Flach J, Porret N, Joncourt R, Wiedemann G, et al. Challenges in the introduction of next-generation sequencing (NGS) for diagnostics of myeloid malignancies into clinical routine use. Blood Cancer J. 2018;8(11):113. doi:10.1038/s41408-018-0148-6

8. Isaic A, Motofelea N, Hoinoiu T, Motofelea AC, Leancu IC, Stan E, et al. Next-Generation Sequencing. Rev Its Transform Impact Cancer Diagn Treat Resist Manag Diagn. 2025;15(19):2425. doi:10.3390/diagnostics15192425

9. Xie W, Jiang X, Huang J, Qin M, Bi Z. Research advances in the adjunctive diagnosis of acute myeloid leukemia. Front Oncol. 2025;15:1634935. doi:10.3389/fonc.2025.1634935

10. Dalal BI, Mansoor S, Manna M, Pi S, Sauro GD, Hogge DE. Detection of CD34, TdT, CD56, CD2, CD4, and CD14 by Flow Cytometry Is Associated With NPM1 and FLT3 Mutation Status in Cytogenetically Normal Acute Myeloid Leukemia. Clin Lymphoma Myeloma Leuk. 2012;12(4):274–9. doi:10.1016/j.clml.2012.01.003

11. Ehninger A, Kramer M, Röllig C, Thiede C, Bornhäuser M, Bonin M, et al. Distribution and levels of cell surface expression of CD33 and CD123 in acute myeloid leukemia. Blood Cancer J. 2014;4(6):218–218. doi:10.1038/bcj.2014.39

12. González-Guerrero L, Castellet H, Martínez C, González N, Guijarro F, Lloveras N, et al. CD200 in acute myeloid leukemia: marked upregulation in CEBPA biallelic mutated cases. Diagn Pathol. 2025;20(1):56. doi:10.1186/s13000-025-01655-w

13. Liu YR, Zhu HH, Ruan GR, Qin YZ, Shi HX, Lai YY, et al. NPM1-mutated acute myeloid leukemia of monocytic or myeloid origin exhibit distinct immunophenotypes. Leuk Res. 2013;37(7):737–41. doi:10.1016/j.leukres.2013.03.009

14. Lewis JE, Cooper LAD, Jaye DL, Pozdnyakova O. Automated Deep Learning-Based Diagnosis and Molecular Characterization of Acute Myeloid Leukemia Using Flow Cytometry. Mod Pathol. 2024 Jan 1;37(1):100373. doi:10.1016/j.modpat.2023.100373

15. Hu Z, Tang A, Singh J, Bhattacharya S, Butte AJ. A robust and interpretable end-to-end deep learning model for cytometry data. Proc Natl Acad Sci U S A. 2020;117(35):21373. doi:10.1073/pnas.2003026117

16. Ojala M, Garriga GC. Permutation Tests for Studying Classifier Performance. In: 2009 Ninth IEEE International Conference on Data Mining [Internet]. 2009. p. 908–13. Available from: https://doi.org/10.1109/ICDM.2009.108 doi:10.1109/ICDM.2009.108

17. Ellis B, Haaland P, Hahne F, Meur N, Gopalakrishnan N, Spidlen J, et al. flowCore: flowCore: Basic structures for flow cytometry data (Version 12804). [Internet]. 2026. Available from: https://bioconductor.org/packages/flowCore

18. Emmaneel A, Quintelier K, Sichien D, Rybakowska P, Marañón C, Alarcón-Riquelme ME, et al. PeacoQC: Peak-based selection of high quality cytometry data. Cytometry. 2022;101(4):325–38. doi:10.1002/cyto.a.24501

19. Pedregosa F, Varoquaux G, Gramfort A, Michel V, Thirion B, Grisel O, et al. Scikit-learn: Machine Learning in {P}ython. Vol. 12. 2011:2825--2830.

20. Eckardt JN, Middeke JM, Riechert S, Schmittmann T, Sulaiman AS, Kramer M, et al. Deep learning detects acute myeloid leukemia and predicts NPM1 mutation status from bone marrow smears. Leukemia. 2022;36(1):111–8. doi:10.1038/s41375-021-01408-w

21. Kockwelp J, Thiele S, Bartsch J, Haalck L, Gromoll J, Schlatt S, et al. Deep learning predicts therapy-relevant genetics in acute myeloid leukemia from Pappenheim-stained bone marrow smears. Blood Adv. 2023;8(1):70–9. doi:10.1182/bloodadvances.2023011076

22. Grinsztajn L, Oyallon E, Varoquaux G, Grinsztajn L, Oyallon E, Varoquaux G. Why do tree-based models still outperform deep learning on typical tabular data? Adv Neural Inf Process Syst. 2022;35:507–20.

23. Couckuyt A, Gassen S, Emmaneel A, Janda V, Buysse M, Moors I, et al. Unraveling genotype–phenotype associations and predictive modeling of outcome in acute myeloid leukemia. Cytometry B Clin Cytom. 2025;108(5):366--377. doi:10.1002/cyto.b.22230

24. Talle MA, Rao PE, Westberg E, Allegar N, Makowski M, Mittler RS, et al. Patterns of antigenic expression on human monocytes as defined by monoclonal antibodies. Cell Immunol. 1983;78(1):83–99. doi:10.1016/0008-8749(83)90262-9

25. Robles EE, Jin Y, Smyth P, Scheuermann RH, Bui JD, Wang HY, et al. A cell-level discriminative neural network model for diagnosis of blood cancers. Bioinformatics. 2023;39(10):585. doi:10.1093/bioinformatics/btad585

26. Gao Q, Chan A, Zhang J, Sun X, Burke A, Miu O, et al. 28-color single tube for flow cytometric assessment of myeloid maturation, myeloid neoplasia, and acute myeloid leukemia minimal/measurable residual disease. Cytometry B Clin Cytom. 2025;108(3):198–211. doi:10.1002/cyto.b.22233